\documentclass{bmvc2k}

\title{Functional Hand Type Prior for 3D Hand Pose Estimation and Action Recognition \\ from Egocentric View Monocular Videos}

\addauthor{Wonseok Roh}{paulroh@korea.ac.kr}{1}
\addauthor{Seung Hyun Lee}{easter3163@korea.ac.kr}{1}
\addauthor{Won Jeong Ryoo}{petac@korea.ac.kr}{1}
\addauthor{Jakyung Lee}{2023020917@korea.ac.kr}{1}
\addauthor{Gyeongrok Oh}{dhrudfhr98@korea.ac.kr}{1}
\addauthor{Sooyeon Hwang}{hsy506@korea.ac.kr}{1}
\addauthor{Hyung-gun Chi}{chi45@purdue.edu}{2}
\addauthor{Sangpil Kim}{spk7@korea.ac.kr}{1$^\ast$}

\addinstitution{
 Department of Artificial Intelligence,\\
 Korea University,\\
 Seoul, Republic of Korea
}
\addinstitution{
 Electrical and Computer Engineering,\\
 Purdue University,\\
 West Lafayette, Indiana, USA
}

\runninghead{W.Roh et al.}{Functional Hand Type Prior}

\def\eg{\emph{e.g}\bmvaOneDot}

\def\etal{\emph{et al}\bmvaOneDot}
\newcommand{\myparagraph}[1]{\vspace{2pt}\noindent{\bf #1}}
\usepackage{wrapfig}
\usepackage{tikz}
\usepackage{comment}
\usepackage{amsmath,amssymb}
\usepackage{color}
\usepackage{booktabs}
\usepackage{adjustbox}
\usepackage[accsupp]{axessibility}
\usepackage{multirow}
\usepackage{array}
\usepackage{mathtools}
\usepackage{adjustbox}
\usepackage{float}
\usepackage{arydshln}
\begin{document}

\maketitle
\vspace{-1.5em}
\begin{abstract}

Current methods for egocentric view action recognition often face challenges in perceiving dynamic hand movements relying solely on geometrical or physical information.  
In this work, we effectively address this problem by gaining insights into the correlation between functional hand configurations and objects, which improves the detailed interpretation of real-world scenarios.
To this end, we introduce a practical taxonomy of hand types based on the functioning perspective and utilize it for per-frame hand type labeling on existing datasets. We also propose a novel hand action recognition framework considering semantic details of the hand type as prior. This approach boosts the network's understanding of the continuous hand interaction throughout the action sequence. 
Our whole pipeline consists of three main modules: (1) Feature Extraction, (2) Egocentric Knowledge Module, which estimates 3D hand pose, object category, and hand type leveraging short-term cues, and (2) Egocentric Action Module, which aggregates per-frame knowledge, including text embeddings of hand type, over a longer time. In our extensive experiments with large-scale benchmarks, FPHA and H2O, our model outperforms current state-of-the-art methods, demonstrating its superior performance.

\end{abstract}


\vspace{0.5em}
\section{Introduction}
\label{sec:intro}
\vspace{-0.5em}
Understanding dynamic interacting human hands is a promising computer vision task because people use their hands to handle objects and communicate with others throughout daily activities. Recently, various hand-based visual applications have been proposed, such as human-robot collaboration~\cite{wang2018controlling, mazhar2019real, gao2020robust, gao2021dynamic} and imitation learning~\cite{qin2022one, qin2022dexmv, mandikal2022dexvip}. Moreover, the advancement of low-cost wearable sensors and VR/AR technologies motivates the computer vision community to tackle egocentric view hand action recognition. 



\begin{figure}[t]
    \centering
    \includegraphics[width=.95\textwidth]{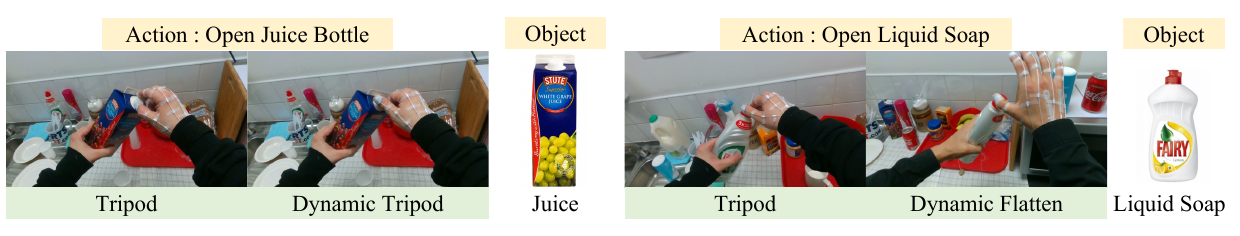}
    \vspace{-1.em}
    \caption{Examples of actions and corresponding hand types from the FPHA dataset~\cite{garcia2018first}. While each action shares the goal of opening objects, the specific interactions between hands and each object are different. Further, both activities involve holding and opening the lid, but the semantic functions of these motions differ significantly. To highlight the continuously repeated shape during the operation, we represent it using \textit{``Dynamic''}. 
    }
    \label{fig:1}
\end{figure}

In real-world egocentric view scenarios, substantial occlusion and truncation often occur, especially when the hand actively interacts with other hands or objects. 
To address these issues, recent approaches~\cite{fathi2011understanding, li2015delving, singh2016first, kwon2021h2o, tekin2019h+, yang2020collaborative, wen2022hierarchical, liu2017global, shi2019two, yan2018spatial} in hand action recognition primarily focus on the temporal context of 3D hand position or high-level object labels.
However, these methods still face challenges in perceiving dynamic hand movements relying solely on geometrical information.
In this work, we go beyond mere physical information and focus on the semantical hand interactions to provide valuable details for a comprehensive understanding of hand action sequences.
For example, as illustrated in Fig.~\ref{fig:1}, there are semantic differences between everyday hand actions such as~\textit{``Open Juice Bottle''} and~\textit{``Open Liquid Soap''}. 
Although each action shares the goal of opening objects, the specific interactions between hands and each object differ.
Therefore, gaining insights into the semantic relationships between functional hand configurations and other hands or various objects is essential for addressing complex real-world hand scenarios.
From this observation, we consider utilizing contextual knowledge of hand types to understand complex hand interactions above simply relying on the physical hand positions.

One of the best ways to specify hand type at semantic level is hand type taxonomy. 
The concept of hand type represents a figurative expression of human intention when conducting tasks with hands.
Most previous studies~\cite{feix2015grasp, schlesinger1919mechanische, napier1956prehensile, cutkosky1989grasp, feix2009comprehensive, bullock2013finding} organize hand type taxonomies based on the properties of objects since humans usually use the similar grasp types for specific objects. In other words, they design and utilize hierarchical hand type categorization criteria based on object grasp manner. Although this hierarchy helps explain the appearance of the hand, this approach overlooks the importance of hand functionality or how the hand operates in various tasks. The function of the hand plays a crucial role in effectively understanding each action, as it accurately reflects the hand-object correlation as well as the intention behind the movement. To this end, we carefully redefine a practical taxonomy of hand types focused on the functioning perspective for diverse hand-related vision tasks, including egocentric view action recognition. 
Then, we supervise the hand action recognition network with hand type labels annotated with newly defined taxonomy.

Built upon \cite{wen2022hierarchical} framework, which considers the high-level action recognition task as a mixture of two low-level tasks: 3D hand pose estimation and object classification, we especially utilize hand type estimation network to provide valuable semantic cues. Our model consists of three main modules: (1) Feature Extraction, (2) Egocentric Knowledge Module, which estimates 3D hand pose, object category, and hand type leveraging short-term cues, and (3) Egocentric Action Module, which aggregates per-frame knowledge, including text embeddings of hand type, with long-term cues. To the best of our knowledge, we are the first to adopt the usefulness of semantic cues of hand type for hand action recognition. Our proposed framework outperforms the existing state-of-the-art works on 3D hand pose estimation and action recognition. To summarize, our main contributions are listed as follows:

\begin{itemize}
 \vspace{-.2em}
 \item We newly define the practical taxonomy of functional hand types based on the functioning perspective for diverse hand-related vision tasks, including hand action recognition. We also provide precise hand type annotations for existing datasets. 
\vspace{-.2em}
 \item We present a novel hand action recognition framework utilizing semantic details of the hand type in each frame. This approach guides the entire network to learn deep understandings of the continuous interaction across hands and objects throughout the action sequence. To the best of our knowledge, our work is the first to leverage the semantic knowledge of hand type in the context of hand perception.
\vspace{-.2em}
 \item 
 We analyze the effectiveness of our proposed hand type prior framework on large-scale benchmarks, including FPHA and H2O. Extensive experiments validate that our method generates a new State-of-the-Art score on 3D hand pose estimation and action recognition from egocentric view monocular videos.
\vspace{-.8em}
\end{itemize}


\vspace{-0.5em}
\section{Related Work}
\label{sec:related}
\vspace{-0.5em}
\textbf{3D Hand Pose Estimation \& Hand Action Recognition}
Hand action typically involves the interaction of hands and objects.
3D hand position provides important information about the object geometry and grasp type, which is positively correlated with hand action~\cite{cai2016understanding, yang2015grasp, cai2017ego, shilkrot132019workinghands}.
Therefore, the hand pose is an influential feature for hand action recognition~\cite{garcia2018first,tekin2019h+}. 
Recently, CNN-based approaches \cite{wang2014action, narasimhaswamy2019contextual}, and graph convolutional network-based methods~\cite{kwon2021h2o, shi2019two, yan2018spatial} have learned important hand pose information from extracted meaningful spatial features.
However, since hand actions are not static representations, it is difficult to perceive the continuous movement of hands.
Thus, it is crucial to consider temporal information to improve understanding of hand activities. Alternative action recognition models such as temporal CNN-based~\cite{ke2017new, luvizon20182d, kim2017interpretable, yang2020collaborative}, LSTM-based~\cite{tekin2019h+, liu2016spatio, liu2017global}, and two-stream networks~\cite{carreira2017quo, feichtenhofer2016convolutional, simonyan2014two, feichtenhofer2019slowfast} appear, yet they still rely on either information.
To consider both spatial and temporal information, current state-of-the-art works~\cite{liu2020forecasting, li2021trear} introduce transformer-based approaches with the multi-head self-attention network that helps find the relationship between the input sequences.
From these observations, we adopt hierarchical transformers model architecture to learn not only spatial information through positional encoding, but also geometric 3D hand position with short-term temporal cues and semantic action flow with long-term temporal cues hierarchically at once.

\vspace{-1em}\noindent\newline\textbf{Hand Type Taxonomy}
The relation between hand grasp type and object has been widely studied for decades~\cite{klatzky1987knowledge, gilster2012contact, cai2016understanding}.
The grasp types reveal the characteristics of the object because people generally use the same or similar hand grasp types for particular things.
Early work by Schelesinger~\etal~\cite{schlesinger1919mechanische} first categorized hand grasp type into six major types based on object shape, hand surfaces, and hand shape. 
Also, Napier~\etal~\cite{napier1956prehensile} divided hand movements into two main groups, prehensile and non-prehensile, and established the concept of power and precision grasp types.
Furthermore, in recent hand-related works~\cite{manti2015novel, juravle2011attention, xiong2016design}, the grasp taxonomy proposed by Feix~\etal~\cite{feix2015grasp} has been widely used in hand analysis.
Understanding the hand grasp type helps the model recognize the user's intention more accurately and improves the accuracy of hand action recognition.
Therefore, we precisely redefine the taxonomy of hand grasp types based on functionality for practical action comprehension.

\vspace{-0.5em}
\section{Function-Based Hand Type Taxonomy}
\label{sec:taxonomy}
\vspace{-0.5em}
\begin{figure}[t]
    \centering
    \includegraphics[width=0.95\textwidth]{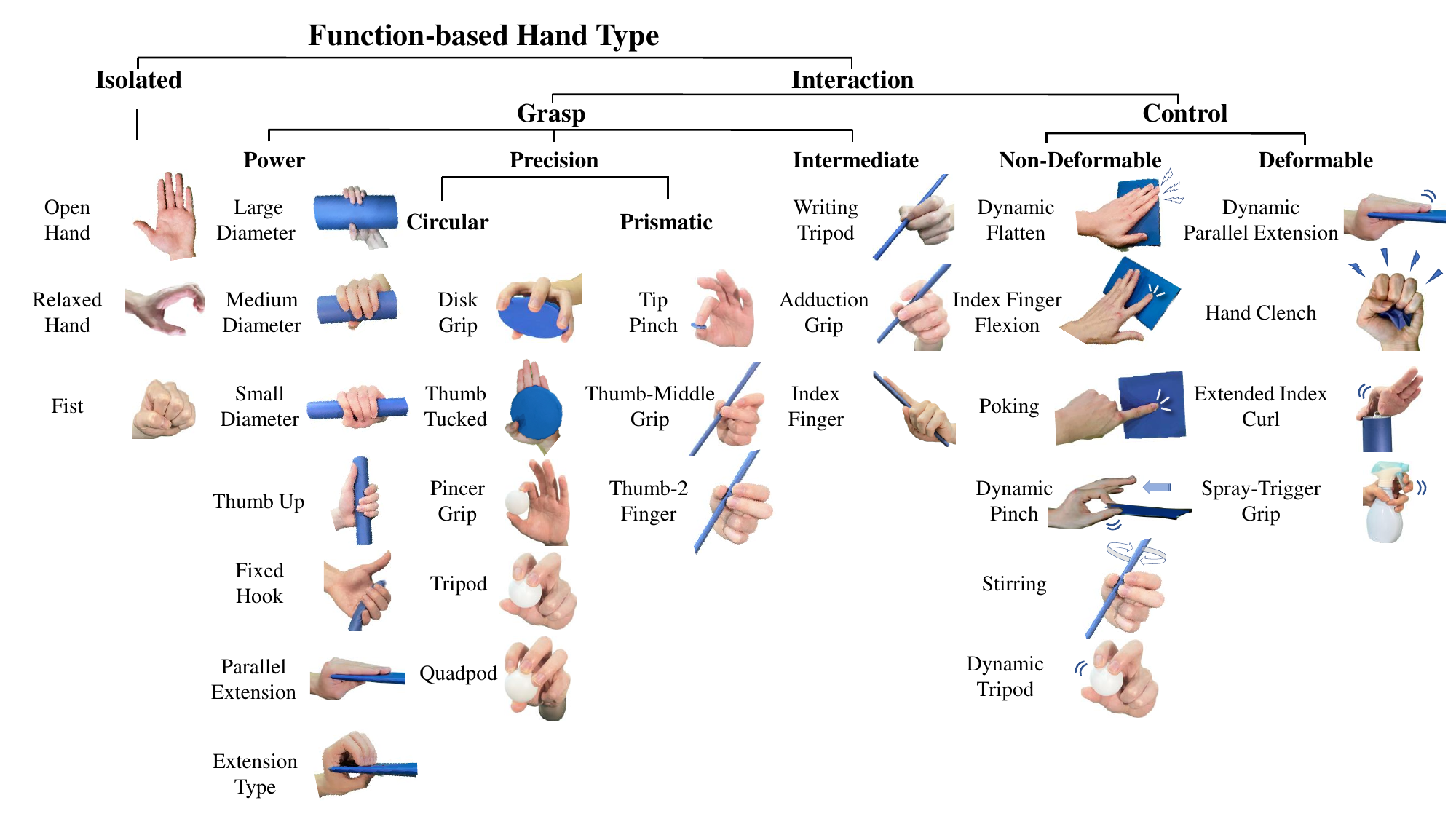}
    \vspace{-1em}
    \caption{Taxonomy of hand types based on functionality. We categorize the mainly used hand types focusing on the role of the hand. Beyond static hand grasp types, we further define dynamic hand types to \textit{Control} to explain more real-life hand behaviors. }
    \label{fig:tax_hand}
    \vspace{-1em}
\end{figure}


\label{sec:hand_taxonomy}

Insight into the context between functional hand configurations and objects could significantly advance understanding of hand action. In this work, we consider the semantic prior of the human-level hand type as a critical indicator of perceiving egocentric view scenes. To achieve this goal, we first carefully design a useful taxonomy of hand types based on the functioning perspective and make use of them for effective dataset curation.

\myparagraph{Redesign Taxonomy} The grasp type is a figurative expression of human intention when performing tasks with hands, and it is closely related to hand action. Furthermore, since we interact with the other person's hand or object to perform many activities, it is very important to identify the hand type in understanding the relationship between the hand and the others. Most previous studies~\cite{cai2016understanding, iberall1997human, bullock2012assessing} have defined hand type taxonomies based on the characteristics of objects interacting with the hand because people usually use the same or similar grasp types for specific objects. They design and utilize hierarchical hand type categorization criteria based on grasp manner. We empirically observe that this hierarchy helps understand the appearance of the hand. However, to better understand the activity in the egocentric view video clip, it is more beneficial to categorize based on the function and role of the hand as well as the appearance of the hand. In other words, there are limitations to traditional taxonomy in explaining various hand action scenarios in everyday life. Our carefully designed taxonomy starts from this observation.

As shown in Fig.~\ref{fig:tax_hand}, we first categorize the mainly used hand types into two groups: \textit{Isolated} and \textit{Interaction}. \textit{Isolated} represents independent hand types that do not interact with other people or objects, and \textit{Interaction} describes hand types that are actively interacting with other people or objects. We then organize the hand types as \textit{Interaction} into \textit{Grasp} and \textit{Control}, focusing on the functions in the activity. \textit{Grasp} includes hand types that hold an object even over time, which are further divided into \textit{Power}, \textit{Precision}, and \textit{Intermediate} depending on the hand's appearance. We also distinguish between \textit{Circular} and \textit{Prismatic} pivoting on the shape of the interacting objects. On the other hand, \textit{Control} includes hand types that manipulate objects over time. Here, we classify hand types, concentrating on the deformation of the object. For example, the hand type that turns the lid does not deform the object, but the hand type that crumples the paper or the hand type that opens the can deform the object. Thus, we differentiate them as \textit{Non-Deformable} and \textit{Deformable} according to these criteria. 

\myparagraph{Dataset Annotation} Ultimately, we perform hand type annotation for all frames of the FPHA~\cite{garcia2018first} and H2O~\cite{kwon2021h2o} dataset based on the newly defined taxonomy. We implement hand type labeling on the right hand for FPHA, as it contains information only for the right hand. However, for the H2O dataset, which provides data for both hands, we label both hands. Note that more details about dataset annotation are provided in supplemental material.



\vspace{-0.5em}
\section{Learning Functional Hand Type}
\vspace{-0.5em}
\label{sec:method}
In this section, we advocate for leveraging human-level hand type knowledge to provide semantically rich cues beyond using simply physical-level hand pose information. In particular, we explicitly guide the action recognition pipeline via rich semantic prior of the hand type based on hand type taxonomy that we proposed in Sec.~\ref{sec:hand_taxonomy}. 

It is noteworthy that temporal information for estimated 3D hand position and high-level object labels enhances action recognition accuracy. Although existing techniques~\cite{tekin2019h+,kwon2021h2o,wen2022hierarchical} benefit from the generous geometric potential of 3D position, they do not cover diverse real-world scenarios. 
Specifically, they overlook the semantic role of hand motions in various scenarios and still face challenges in perceiving dynamic hand movements relying solely on geometrical or physical information.
Therefore, we propose to utilize semantic knowledge of hand type in each frame to encourage the network to understand the continuous interaction between the primary hand and the assistive hand/object in the video clip.

\begin{figure}[t]
    \centering
    \includegraphics[width=\textwidth]{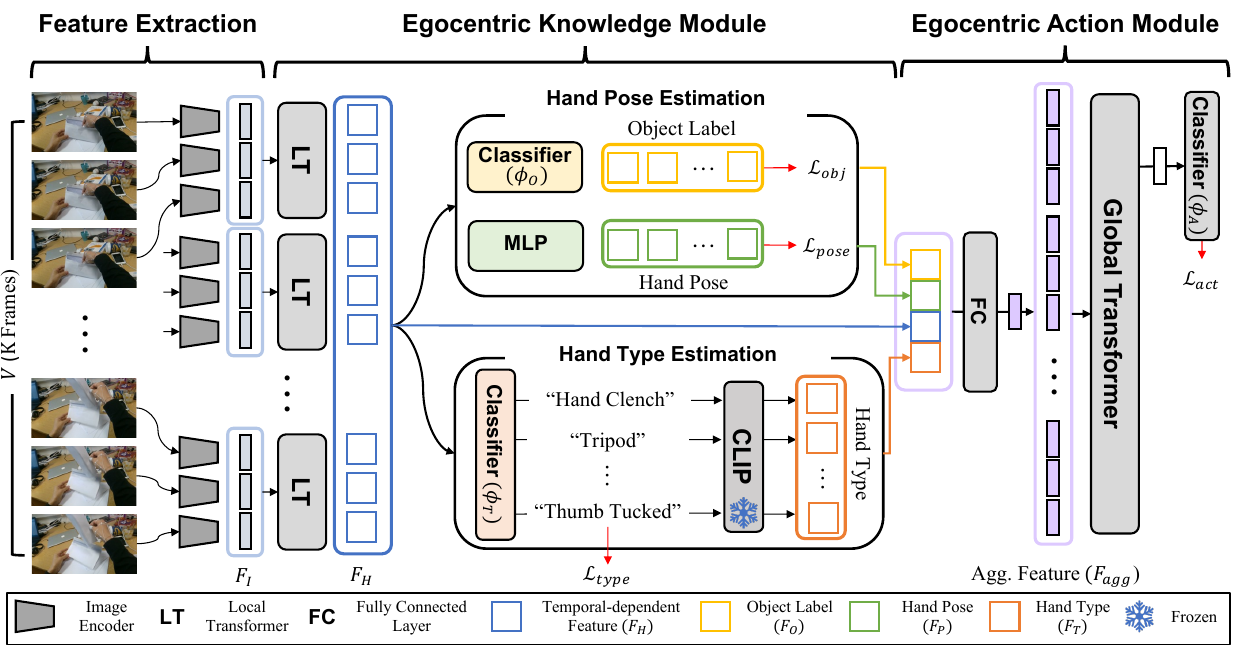}
    \vspace{-1.8em}
    \caption{Overview of our proposed model. Our framework consists of three main modules: (1) Feature Extraction, (2) Egocentric Knowledge Module, which estimates 3D hand pose, object category and hand type leveraging short-term temporal cues, and 
    (3) Egocentric Action Module, which aggregates per-frame pose, object and hand type information over a longer time span. }
    \label{fig:architecture}
    \vspace{-1.0em}
\end{figure}

\vspace{-0.5em}\subsection{Overview}
\label{sec:overview}
We illustrate the outline of our framework in Fig~\ref{fig:architecture}.
Built upon hierarchical temporal transformer (HTT~\cite{wen2022hierarchical}), which considers the high-level action recognition task as a mixture of two low-level tasks: 3D hand pose estimation and object classification, we especially utilize hand type estimation network to provide valuable semantic cues. Our model consists of three main modules: (1) Feature Extraction, (2) Egocentric Knowledge Module, which estimates 3D hand pose, object category and hand type leveraging the short-term temporal cue and (3) Egocentric Action Module, which aggregates per-frame pose, object and hand type information over a longer time span. First, our model takes aligned 2D video clip $V=\{X_{i}\in\mathbb{R}^{3{\times}H{\times}W}|i=1,...,K\}$ consisting of $K$ frames as input, which are converted into feature vector $F_{I}$ containing fine details. Then we employ temporal-dependent features $F_{H}$ from the Local Transformer to estimate the per-frame 3D hand pose with feature vector $F_{P}$, which contains geometric potential, and the category of interacting objects with feature vector $F_{O}$. Additionally, the hand type estimation network takes $F_{H}$ and outputs hand type feature vectors $F_{T}$. Here, we utilize language models to provide the deep semantic cues of hand type based on our proposed functional hand type taxonomy.
Subsequently, the Egocentric Action Module aggregates the predicted embeddings: hand position $F_{P}$, object category $F_{O}$, and hand type $F_{T}$ for action recognition.


\vspace{-0.5em}\subsection{Egocentric Knowledge Module}

To construct Egocentric Knowledge Module input sequence, we first divide the long video clip $V$ into $m$ consecutive segment $\mathtt{seg_k}(V)=(\bar{V}_1,\bar{V}_2,...,\bar{V}_m)$, where $m$ denotes $\lceil K/k \rceil$. In order to capture the temporal cue of consecutive segment for hand pose estimation, the module processes each segment $\bar{V}\in\mathtt{seg_k}(V)$ in parallel. Then transformer takes the sequence of per-frame feature vector $F_{I}$ from the image encoder and outputs temporal-dependent features $F_{H}$. To decode the hand pose information, these features $F_{H}$ are fed into simple MLP layers, yielding joint coordinates in the 2D image plane $P^{2D}\in\mathbb{R}^{J\times{2}}$ and the joint depth to the camera $P^{dep}\in\mathbb{R}^{J\times{1}}$. We train the 3D pose estimation module to minimize the following pose loss (L1-loss):
\begin{equation}\label{eq:hand}
    {\mathcal{L}}_\text{pose}=\frac{1}{J}(||P^{2D}-P^{2D}_{gt}||_1 + \lambda_\text{pose} ||P^{dep}-P^{dep}_{gt}||_1)
\end{equation}
where $J$ denotes hand joints and $\lambda_\text{pose}$ is a hyperparameter to balance the different intensities of the 2D and depth losses. The 3D positions of the hand joints in the camera space $P^{3D}\in\mathbb{R}^{J\times{3}}$ for $I$ can be inferred operating the camera intrinsics. 

In addition to using 3D hand pose information, which gives precise geometric knowledge about egocentric view scenarios, we advocate for leveraging hand type information to guide the entire network to learn deep semantic understanding. In particular, the context details of hand type can serve as a practical semantic key representing hand-object relationships for identifying hand actions. Thus, we introduce a simple but novel hand-type classification network $\phi_T$ to predict the hand type $t_i$ for $i=\{1, 2, \dots, N_t\}$ from temporal-dependent features $F_{H}$. Given the ground truth hand type label $t_{gt}$ from the dataset annotation process (see Sec.~\ref{sec:hand_taxonomy}), the target probability is defined as a one-hot vector $w_t$. For training, we formulate the following cross-entropy loss to train the hand type classification network $\phi_T$:
\begin{equation}
{\mathcal{L}}_{type} = -\mathbb{E}_{t, w_t\sim\mathbb{D}}\Bigg[\sum_{r\in{N_t}}w_t[r] \log \phi_T(t)[r]\Bigg] 
\end{equation}
where $\mathbb{D}$ is (input) data distribution.
Also, we predict the object category $o_i$ in each frame with the object classification network $\phi_O$. Similar to hand type classifier, the classifier $\phi_O$ is supervised to minimize the cross entropy loss ${\mathcal{L}}_{obj}$.

\vspace{-1em}\subsection{Boosting the Action Recognition with Hand Type Prior}
We introduce the strategies using semantic prior of the human-level hand type, potentially improving not only action recognition but also 3D hand pose estimation. Specifically, we explicitly pilot the Egocentric Action Module to learn rich semantic hand type variant information via utilizing the power of the prevalent language model, Contrastive Language-Image Pre-Training~\cite{radford2021learning} (CLIP). Given the candidate of hand type from the hand type classification network $\phi_T$, we map the type number $t_i$ with the corresponding text descriptions (\eg, ~\textit{``Hand Clench''},~\textit{``Tip Pinch''}). Next, the text label of hand type goes through the large-scale pre-trained language model, which outputs text embedding vector $F_{T}$. Importantly, we consider these text embedding vector of predicted hand type descriptions as critical indicators to deliver semantic knowledge to the action recognition pipeline. 

\vspace{-1em}\subsection{Egocentric Action Module}
We adopt the previous approaches~\cite{kenton2019bert, dosovitskiy2020image} presenting trainable tokens to aggregate the global information across the input video clip $V$. Each token encodes short-term temporal information such as temporal-dependent features, hand pose, object label, and hand type details. We design the fully connected layer for each cues, which outputs features of the same dimension, then concatenates these aligned features for input of the Global Transformer as follows:
\begin{equation}
F_{agg} = {\mathtt{FC}[F_{H}~{\oplus}~F_{P}~{\oplus}~F_{O}~{\oplus}~F_{T}]}
\end{equation}
where $\oplus$ indicates channel-wise concatenation of feature vectors and $\mathtt{FC}[\cdot]$ reduces the features into $d$-dim to fit in the token dimension of Global Transformer. After mixing features that potentially contain geometric and semantic knowledge, we feed these feature vector $F_{agg}$ to Global Transformer, which outputs action tokens.
Here, we utilize an action classification head $\phi_A$ to recognize action label $a_i$ for $i=\{1, 2, \dots, N_a\}$ from action tokens. 
For supervision, we formulate the following cross-entropy loss to train the action classification head $\phi_A$:
\begin{equation}
{\mathcal{L}}_{act} = -\mathbb{E}_{a, w_a\sim\mathbb{D}}\Bigg[\sum_{r\in{N_a}}w_a[r] \log \phi_A(a)[r]\Bigg] 
\end{equation}
where $w_a$ denotes one hot-encoded action labels and $\mathbb{D}$ represents (input) data distribution.


\subsection{Training}
Our entire network is trained end-to-end by minimizing the following loss $\mathcal{L}_{\text{total}}$:
\begin{equation}\label{eq:total_loss}
    {\mathcal{L}}_\text{total} = 
    \lambda_\text{act}{\mathcal{L}}_\text{act}+
    \frac{1}{K}\sum_{\bar{V}\in\mathtt{seg_k}(V)}\sum_{I\in\bar{V}}(\lambda_\text{pose}{\mathcal{L}}_\text{pose}
    +\lambda_\text{obj}{\mathcal{L}}_\text{obj}
    +\lambda_\text{type}{\mathcal{L}}_\text{type})
\end{equation}
where $\lambda_\text{act}$, $\lambda_\text{pose}$, $\lambda_\text{obj}$ and $\lambda_\text{type}$ are the hyperparameters of respective loss terms.


\vspace{-0.5em}
\section{Experiments}
\vspace{-0.5em}
\label{sec:experiments}
\begin{table}[h]
    \begin{center}
    \caption{Comparison of our novel hand action recognition framework and the state-of-the-art models on the FPHA~\cite{garcia2018first} and H2O~\cite{kwon2021h2o} dataset. We report the classification accuracy of methods based on RGB videos. Note that the H2O dataset provides additional testing split videos, unlike the FPHA dataset, which provides only training and validation split.}
    \vspace{0.5em}
    \label{table:fpha_val}
    \resizebox{\textwidth}{!}{
    \addtolength{\tabcolsep}{-3pt}    
    \begin{tabular}{lccccccc}\toprule
        \parbox{2cm}{~} & ~Joule-color~\cite{hu2015jointly} & ~Two Stream~\cite{feichtenhofer2016convolutional} & ~H+O~\cite{tekin2019h+} & ~Collaborative~\cite{yang2020collaborative} & ~HTT~\cite{wen2022hierarchical} & ~Trear~\cite{li2021trear} & ~Ours~\\\midrule
        Accuracy~($\uparrow$) & 66.78 & 75.30 & 82.43 & 85.22 & 94.09 & 94.96 & ~\textbf{95.13}\\\bottomrule
    \end{tabular}}
    
    \vspace{0.5em}\centering\small {(a) Action recognition accuracy (\%) on FPHA.}
    \vspace{1.0em}
    
    \resizebox{\textwidth}{!}{
    \addtolength{\tabcolsep}{-3pt}    
    \begin{tabular}{lcccccccc}\toprule
        \parbox{2cm}{~} & ~C2D~\cite{wang2018non} & ~I3D~\cite{vadisaction} & ~SlowFast~\cite{feichtenhofer2019slowfast} & ~H+O~\cite{tekin2019h+} & ~ST-GCN~\cite{yan2018spatial} & ~TA-GCN~\cite{kwon2021h2o} & ~HTT~\cite{wen2022hierarchical} & ~Ours~\\\midrule
        Val Accuracy~($\uparrow$) & 76.10 & 85.15 & 86.00 & 80.49 & 83.47 & 86.78 & 90.16 & ~\textbf{91.80}\\\midrule
        Test Accuracy~($\uparrow$) & 70.66 & 75.21 & 77.69 & 68.88 & 73.86 & 79.25 & 86.36 & ~\textbf{89.67}\\\bottomrule
    \end{tabular}}
    
    \vspace{0.5em}\centering\small {(b) Action recognition accuracy (\%) on  H2O.}
    \vspace{-2.5em}
\end{center}
\end{table}

\subsection{Experimental Setups}

\myparagraph{Dataset}
We train and evaluate overall performance on two landmark datasets for action recognition from egocentric views: FPHA~\cite{garcia2018first} and H2O~\cite{kwon2021h2o}. These two datasets are collected in various indoor settings and have a frame rate of 30 frames per second. Both datasets provide the ground truth labels for hand pose, action, and object category, and we utilize them for supervision and evaluation. In this work, we annotate the hand type of all frames in these two datasets based on the newly defined taxonomy and use them for training. Note that detailed descriptions of each dataset are provided in the supplemental material.

\myparagraph{Evaluation metrics} To evaluate hand action recognition, we follow the official evaluation protocol of the hand action recognition task. We report classification accuracy over the validation and test split by comparing each video's predicted and ground truth action categories. Also, we evaluate the 3D pose estimation performance in the camera space and the root-aligned (RA) space, which aligns the estimated wrist position with the ground truth for each frame. We report the Percentage of Correct Keypoints (PCK) for joints~\cite{zimmermann2017learning} against different error thresholds and the corresponding Area Under the Curve (AUC). We also utilize Mean End-Point Error (MEPE) metrics for hands~\cite{zimmermann2017learning} in the camera and root-aligned space following HTT~\cite{wen2022hierarchical}. We provide all implementation details in the supplemental material.

\vspace{-0.5em}
\subsection{Comparison with the State-of-the-Arts} 
\vspace{-0.5em}
\myparagraph{Hand Action Recognition}
We compare our proposed method with existing state-of-the-art methods including Joule-color~\cite{hu2015jointly}, Two Stream~\cite{feichtenhofer2016convolutional}, H+O~\cite{tekin2019h+}, Collaborative~\cite{yang2020collaborative}, HTT~\cite{wen2022hierarchical}, and Trear~\cite{li2021trear} on the FPHA~\cite{garcia2018first} dataset (see Table~\ref{table:fpha_val} (a)). On the H2O dataset, we compare ours with C2D~\cite{wang2018non}, I3D~\cite{vadisaction}, SlowFast~\cite{feichtenhofer2019slowfast}, H+O~\cite{tekin2019h+}, ST-GCN~\cite{yan2018spatial}, TA-GCN~\cite{kwon2021h2o} and HTT~\cite{wen2022hierarchical} (see Table~\ref{table:fpha_val} (b)). As reported in Table~\ref{table:fpha_val} (a) and (b), our method generally outperforms other methods with state-of-the-art accuracy~(FPHA~\cite{garcia2018first}: $\mathbf{95.13\%}$, H2O~\cite{kwon2021h2o}: $\mathbf{89.67\%}$). These results emphasize the effectiveness of our method in understanding the interaction between hands and objects.

\myparagraph{Hand Pose Estimation} We also demonstrate considerable performance on 3D hand pose estimation, scoring AUC for 3D PCK at error thresholds ranging from 0 to 50 $mm$ and the MEPE measured in $mm$ within the Root-Aligned (RA) space.
Our method estimates hand pose more precisely than the current state-of-the-art method~\cite{wen2022hierarchical}~(see Table~\ref{table:h2o_test}). These experimental results verify that the semantic knowledge of the hand type benefits not only the action recognition network but also the entire network, resulting in high precision in 3D hand pose estimation. 

\vspace{-1.0em}
\subsection{Ablation Study} 
\vspace{-0.5em}
\myparagraph{Effect of Text-based Hand Type Feature}
In this section, we evaluate the variants of our method across hand type features. As shown in Table~\ref{tab:ablation} (a), we investigate the usage of the hand type feature. The Egocentric Action Module (EAM) performs better when the predicted probability distribution feature vector from the hand type estimation network is used as input than when the hand type is simply predicted as the auxiliary network. Further, utilizing text embedding of hand type improves accuracy over using the predicted feature vector.



\myparagraph{Effect of Hand Type Cue} With four types of cues (image feature, hand pose, object, and hand type), we analyze the effect of each cue on action recognition with and without each cue as input of the egocentric action module. Specifically, on the FPHA~\cite{garcia2018first} and the H2O~\cite{kwon2021h2o} datasets, hand type cue effectively enhances accuracy, as shown in Table~\ref{tab:ablation} (b). We finally observe that employing hand type as a semantic prior plays a key role in action recognition.

\vspace{-0.5em}
\begin{table*}[t]
\caption{
    3D pose estimation performance in Root-Aligned space on the FPHA~\cite{garcia2018first} and H2O~\cite{kwon2021h2o}. We report AUC-RA for 3D PCK-RA at error thresholds ranging from 0 to 50 $mm$ and the MEPE-RA in the unit of $mm$.
    }
    \begin{minipage}[t]{0.5\textwidth}
        \label{table:h2o_val}
        \vspace{-0.5em}
        \resizebox{0.9\textwidth}{!}{%
        \begin{tabular}{lcc}\toprule
                Model & AUC-RA(0-50)~($\uparrow$) & MEPE-RA~($\downarrow$)\\\midrule
                HTT~\cite{wen2022hierarchical} & 0.763 & 12.13 \\\midrule
                Ours & \textbf{0.769} & \textbf{11.79} \\ 
                \bottomrule
                \vspace{0.05em}
        \end{tabular}}
        \vspace{-0.5em}
        \centering\small \\
        {(a) FPHA Dataset}
    \end{minipage}
    \hfill
    \begin{minipage}[t]{0.5\textwidth}
        \label{table:h2o_test}
        \vspace{-0.5em}
        \resizebox{0.9\textwidth}{!}{%
        \begin{tabular}{lcccc}\toprule
                \multirow{2}{*}{Model} & \multicolumn{2}{c}{AUC-RA(0-50)~($\uparrow$)} & \multicolumn{2}{c}{MEPE-RA~($\downarrow$)}\\\cmidrule{2-5}
                ~ & Left & Right & Left & Right \\\midrule
                HTT~\cite{wen2022hierarchical} & 0.674 & 0.648 & 16.59 & 17.91\\\midrule
                Ours & \textbf{0.686} & \textbf{0.662} & \textbf{15.96} & \textbf{17.08} \\ 
                \bottomrule
                \vspace{0.05em}
        \end{tabular}}
        \vspace{-0.5em}
        \centering\small \\
        (b) H2O Dataset
    \end{minipage}
    \vspace{-1em}
\end{table*}

\begin{table}[t]
        \caption{Ablative study of input features for Egocentric Action Module (EAM) on Hand Action Recognition Accuracy (\%). We investigate the usage of the hand type feature in (a). Also, we analyze the effectiveness of each cue on the action recognition task in (b).
        }
        \begin{minipage}[t]{0.5\textwidth}
	    \label{tab:ablation}
    	\resizebox{\textwidth}{!}{%
        \addtolength{\tabcolsep}{-3pt}    
    	\begin{tabular}{ccccc} \toprule
            \multirow{2}{*}{Hand Type} & \multirow{2}{*}{EAM Input} & \multirow{2}{*}{Text Embedding} & \multicolumn{2}{c}{Accuracy~($\uparrow$)}  \\\cmidrule{4-5}
             & & & FPHA~\cite{garcia2018first} & H2O~\cite{kwon2021h2o} \\\midrule
            $\checkmark$ & - & -  & 93.74 & 85.95 \\
            $\checkmark$ & $\checkmark$ & -  & 94.26 & 87.60 \\
            $\checkmark$ & $\checkmark$ & $\checkmark$ &  \textbf{95.13} & \textbf{89.67} \\
           \bottomrule
        \end{tabular}}
        \addtolength{\tabcolsep}{2pt}    
        \centering\small \\
        \vspace{0.5em}
        (a) Effect of Text-based Hand Type Feature 
        \end{minipage}
        \hfill
        \begin{minipage}[t]{0.5\textwidth}
	    \label{tab:ablation}
    	\resizebox{\textwidth}{!}{%
        \addtolength{\tabcolsep}{-3pt}    
    	\begin{tabular}{cccccc} \toprule
            \multicolumn{4}{c}{Input Feature for Egocentric Action Module} & \multicolumn{2}{c}{Accuracy~($\uparrow$)}  \\\cmidrule{5-6}
            Image Feature & Hand Pose & Object Label & Hand Type & FPHA~\cite{garcia2018first} & H2O~\cite{kwon2021h2o} \\\midrule
            $\checkmark$ & $\checkmark$ & $\checkmark$ & - & 94.09 & 86.36 \\
            $\checkmark$ & $\checkmark$ & - & $\checkmark$ & 93.74 & 87.19 \\
            $\checkmark$ & - & $\checkmark$ & $\checkmark$ & 94.61 & 86.78 \\
            $\checkmark$ & $\checkmark$ & $\checkmark$ & $\checkmark$ & \textbf{95.13} & \textbf{89.67} \\
           \bottomrule
        \end{tabular}}
        \centering\small \\
        \vspace{0.5em}
        (b) Effect of Hand Type Cue
        \end{minipage}
\end{table}

\subsection{Qualitative Analysis}
\begin{figure}[t]
    \centering
    \includegraphics[width=0.99\textwidth]{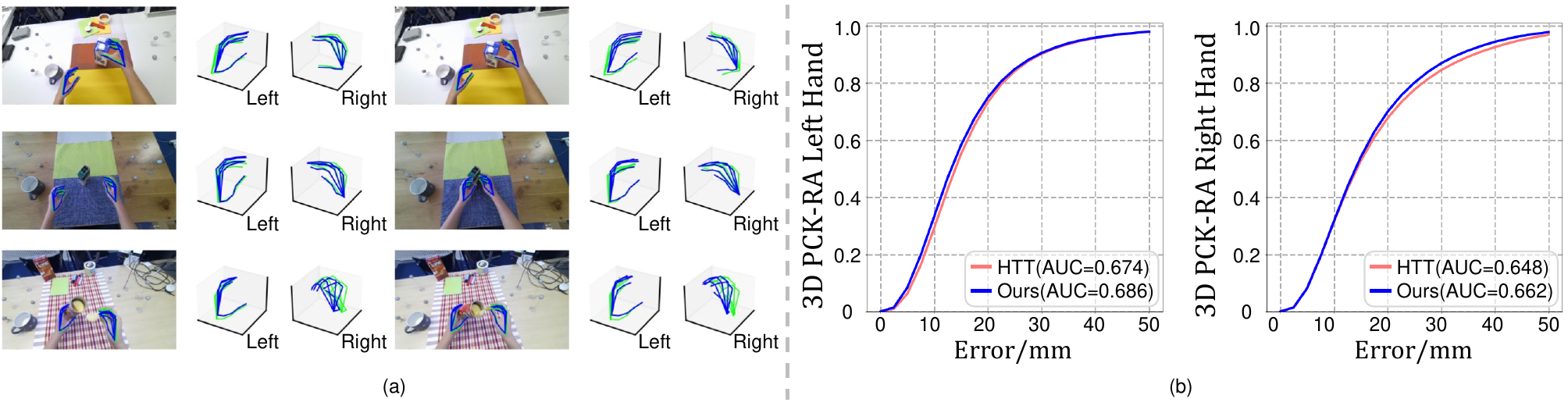}
    \vspace{-1.0em}
    \caption{Qualitative result of our experiments. In (a), the green and blue line represents ground truth and estimated 3D hand pose, respectively. (b) shows the 3D PCK of hand pose estimation results on H2O~\cite{kwon2021h2o} in Root-Aligned space. The blue line indicates the performance of our model, whereas the red line represents the HTT~\cite{wen2022hierarchical}.}
    \label{fig:qualitative}
\end{figure}

In this section, we qualitatively verify the usefulness of our novel framework. Fig.~\ref{fig:qualitative} shows the visualized results of our experiments. In Fig.~\ref{fig:qualitative} (a), the ground truth hand pose (see green lines) and our estimated hand pose (see blue lines) are projected in both 3D space and images. Ours show comparable results compared to ground truth. We also provide 3D PCK-RA graphs of left and right hand pose estimation results on the H2O test split of ours~(see blue lines) vs. HTT~\cite{wen2022hierarchical} (see red lines) in Fig.~\ref{fig:qualitative} (b). This graph validates that our method generates reasonable results in both hands. Overall, our model is robust for estimating hand pose in 3D space. We provide more qualitative results and analysis in supplementary materials.

\vspace{-0.5em}
\section{Conclusion}
\label{sec:conclusion}
\vspace{-0.5em}
In this paper, we present a novel method applying the knowledge of hand type for hand action recognition based on the temporal transformer. This is the first attempt to regard the semantic details of the hand type as a critical indicator for enhancing the perception of egocentric view hand actions. To utilize the knowledge of hand type, we newly define the taxonomy based on hand functionality and annotate hand types for existing large-scale benchmarks. The experiments demonstrate the outstanding performance of our proposed approach.


\vspace{-0.5em}
\section*{Acknowledgement}
\label{sec:acknowledgement}
This work was supported by Electronics and Telecommunications Research Institute~(ETRI) grant funded by the Korean government~(23ZH1300, Research on hyper-realistic interaction technology for five senses and emotional experience, 95\%), Institute of Information \& communications Technology Planning \& Evaluation~(IITP) grant funded by the Korea government~(MSIT)~(No.2019-0-00079, Artificial Intelligence Graduate School Program~(Korea University), 3\%), the National Research
Foundation of Korea grant~(NRF-2022R1F1A10\\74334, 2\%), and results of a study on the "Leaders in Industry-university Cooperation 3.0" Project, supported by the Ministry of Education and National Research Foundation of Korea. This work was supported by Artificial intelligence industrial convergence cluster development project funded by the Ministry of Science and ICT(MSIT, Korea)\& Gwangju Metropolitan City.
\vspace{-0.5em}


\clearpage
\bibliography{egbib}


\end{document}


\maketitle
\vspace{-2.6em}
\section*{Overview}
\label{sec:overview}
\vspace{-0.6em}
In this supplementary material, we supply further explanations and visualizations of our main paper, \textbf{"Functional Hand Type Prior for 3D Hand Pose Estimation and Action Recognition from Egocentric View Monocular Videos"}. 
We first provide additional experimental results to validate our proposed framework (Section~\hyperref[sec:a]{A}).
We also describe implementation details for training (Section~\hyperref[sec:b]{B}).
Next, we explain elaborate descriptions of hand type annotation process (Section~\hyperref[sec:c]{C}).
We further provide more details for large-scale datasets (FPHA~\cite{garcia2018first} and H2O~\cite{kwon2021h2o}) and analysis of hand type distributions based on functional hand type taxonomy (Section~\hyperref[sec:d]{D}).
Moreover, we supply extra details on the following project website:~\url{https://kuai-lab.github.io/bmvc2023fhtp}.



\section*{A. Additional Experimental Results}
\vspace{-0.4em}
\label{sec:a}
\label{sec:experiments}

\begin{figure}[h]
\vspace{-0.5em}
    \centering
    \includegraphics[width=1\textwidth]{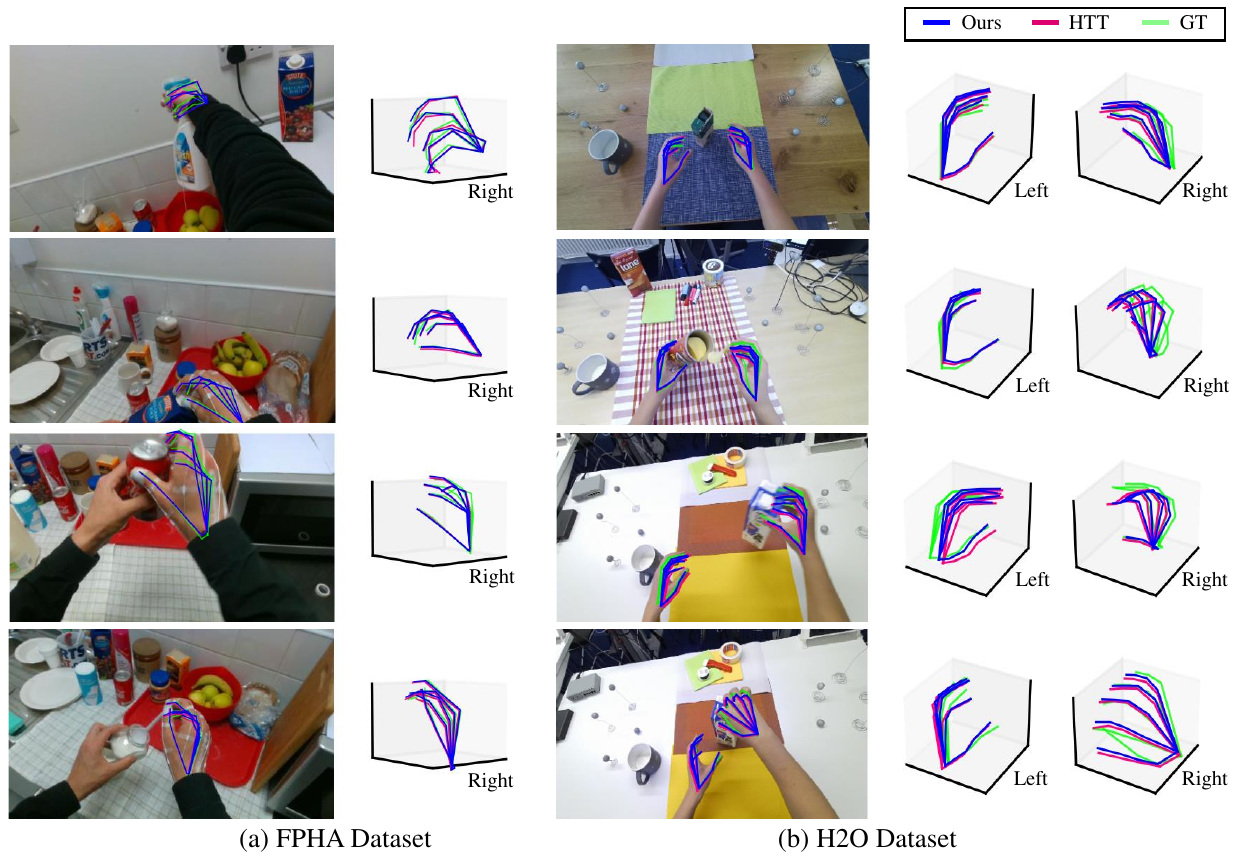}
    \vspace{-1.em}
    \caption{Qualitative comparison of 3D hand pose estimation results between our model~(blue lines) and HTT~(magenta lines) on the (a) FPHA~\cite{garcia2018first} and (b) H2O~\cite{kwon2021h2o} dataset. Both ground truth (green lines) and estimated hand poses are visualized in 3D space and projected to the corresponding 2D image frames. Note that the H2O provides labels for both hands, while the FPHA only includes information for the right hand.}
    \vspace{-2.0em}
    \label{fig:qualitative_result}
\end{figure}

\myparagraph{Qualitative Results} 
In this section, we provide additional qualitative examples and analysis of our proposed model.
In Fig.~\ref{fig:qualitative_result}, we show various visualized results of the 3D hand poses predicted by ours (see blue lines) and HTT~\cite{wen2022hierarchical} (see magenta lines) in 3D space and their projection to corresponding 2D image frames, compared with ground truth (see green lines). We observe that for both FPHA~\cite{garcia2018first} (a) and H2O~\cite{kwon2021h2o} (b), the gap between ours and ground truth is smaller than that between HTT and ground truth. These visualizations qualitatively verify that our model outperforms the baseline model by using functional hand type as semantic prior. In other words, utilizing hand type assists estimating hand pose in 3D space.

\section*{B. Implementation Details}
\label{sec:b}
\label{sec:implementation_datails}

For training, we use two RTX 3090 GPUs with a batch size of 4, allowing for efficient processing and improved performance.
Since the model converged at around 45 epochs, we trained it for a total of 45 epochs with a learning rate $3 \times 10^{-5}$, which is decreased 10 times every 15 epochs.
We applied the Adam optimizer to train the dataset.
Using PyTorch, we implement our experimental setup with the following settings.
Our model architecture adopts $K$ RGB frames, resized to 480 $\times$ 270, as the input sequence.
All $K$ input sequence images are sliced into $k$ segments and fed into ResNet-18 pre-trained on ImageNet to extract image features. Then transformer block takes these features and outputs temporal-dependent features.
We set the dimension size of all features and tokens to 512.
Moreover, to optimize our model efficiently, we apply data augmentation strategies following the previous method~\cite{wen2022hierarchical}.


\section*{C. Annotation Details}
\label{sec:c}
\label{sec:annotation}
\begin{figure}[t]
\centering
\includegraphics[width=\textwidth]{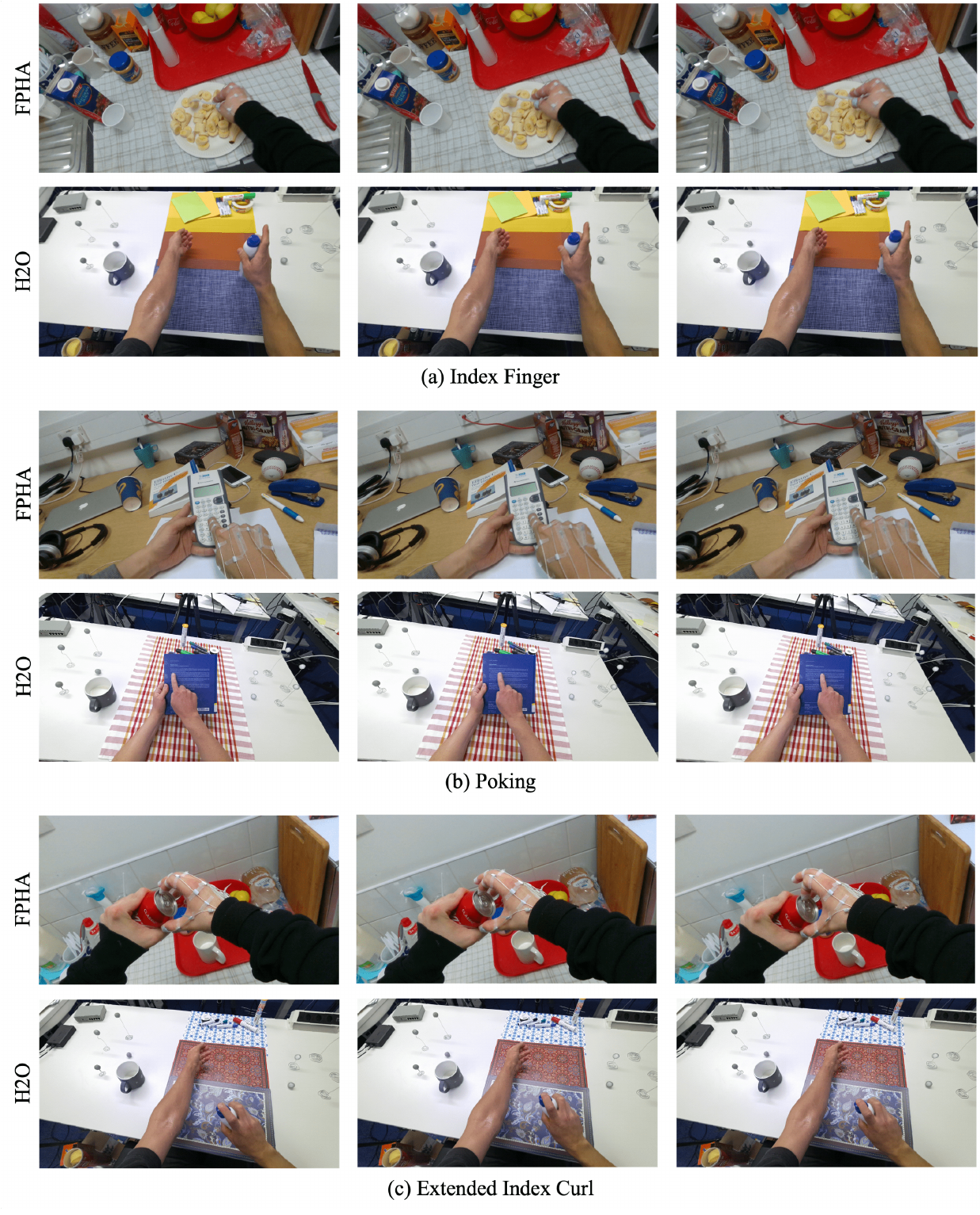}
\vspace{-1.8em}
\caption{Examples of action scenes in which the variation of hand pose is similar, but the action labels are different according to the temporal context of hand function and object types. (a) \textit{``Index Finger''}, (b) \textit{``Poking''}, and (c) \textit{``Extended Index Curl''} are all hand types that use index fingers, but their functions vary depending on the action and object they interact with. \textit{``Index Finger''} represents holding an object while extending the index finger, \textit{``Poking''} represents poking into an object without deforming it, and \textit{``Extended Index Curl''} represents utilizing the index finger to deform an object by applying pressure on its tip.}
\label{fig:handtype3}
\end{figure}

Beyond traditional hand grasp types, we categorize hand types based on functionality. In addition, we newly define additional hand types to describe more real-life hand activities and utilize deep semantic knowledge of hand types.
For example, classic taxonomy includes the \textit{``Index Finger''} category for holding an object while extending the index finger. 
However, as shown in Fig.~\ref{fig:handtype3}, we add two more categories for extending the index finger; \textit{``Poking''} which indicates to poking into an object without deforming it, and \textit{``Extended Index Curl''}, which represents a hand type that utilizes the index finger to deform an object by applying pressure on its tip. Here, we explain the details of our annotation process based on these function-based taxonomy.

We manually annotate hand type all frames of landmark datasets, considering both the appearance of the hand and the various functions the hand performs.
To make consistent annotation, we inspect which hand type certainly appears in specific actions before annotating each frame individually.
Explanation in more detail; for the \textit{``Open Juice Bottle''} action of the FPHA~\cite{garcia2018first} dataset (left side of Fig.~1 in the main paper), we label \textit{``Tripod''} as holding the juice lid and \textit{``Dynamic Tripod''} as opening the lid.
Therefore, the frames just holding the lid are marked as \textit{``Tripod''} and from the frames that start to open, they are marked as \textit{``Dynamic Tripod''} since the hand type changes to manipulate operation. More specifically, when the thumb finger is out while opening the lid, the hand type converts to \textit{``Thumb Up''}.
If the hand is out of frames or occluded by the object, it is annotated as \textit{``Invisible''}.
Also, for the challenging and unclear hand type, we go over whole videos implementing the same action by other subjects and then annotate as the most closely matched hand type.

\begin{figure}[t]
\centering
\includegraphics[width=\textwidth]{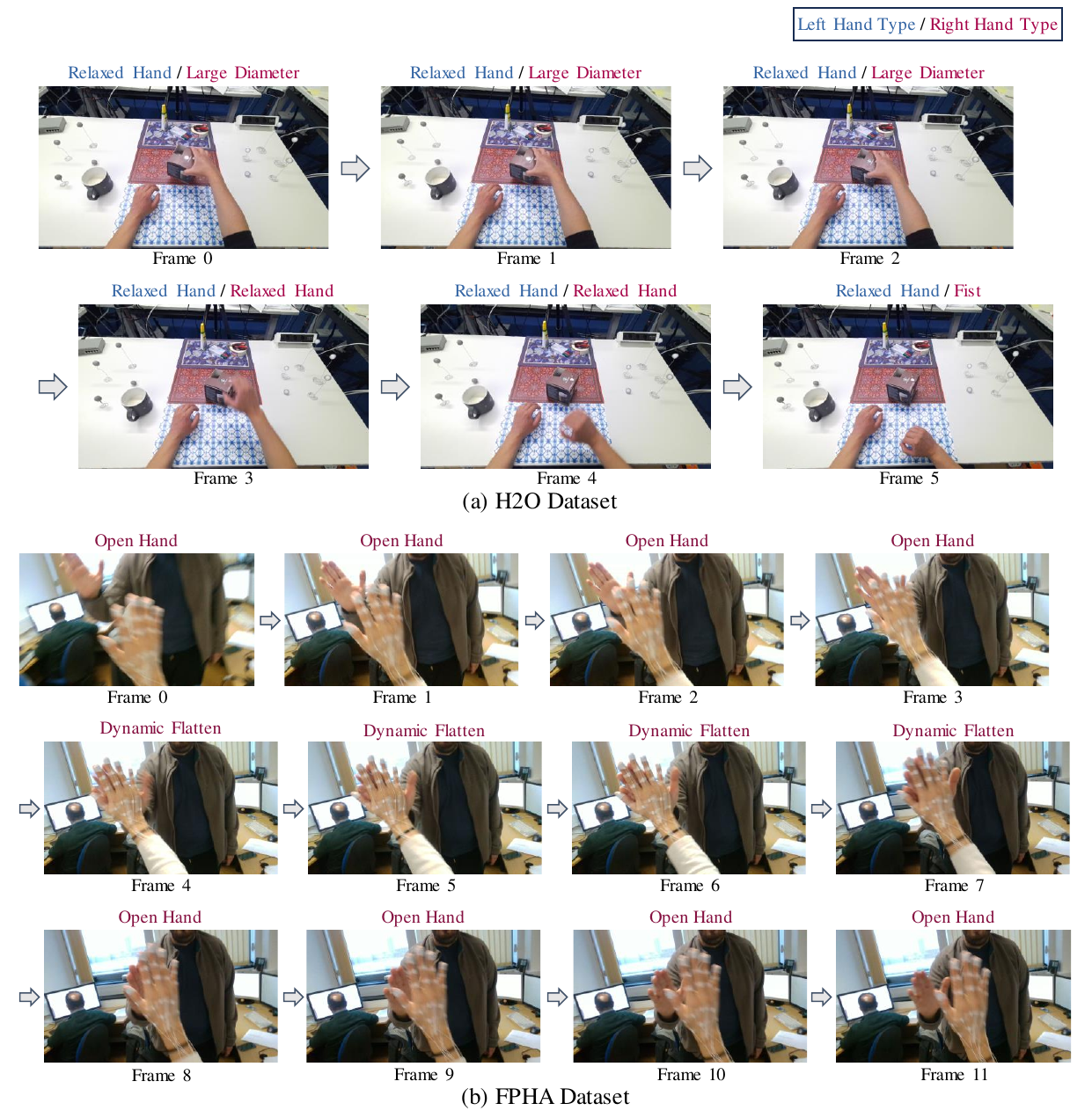}
\vspace{-1.8em}
\caption{
Examples of continuous video frames with hand type annotation. The hand type of each frame changes over time in the video sequence. In (a) and (b), we show frames of the \textit{``Place Cappuccino''} action (H2O~\cite{kwon2021h2o}) in which the hand interacts with the box and the \textit{``High Five''} action (FPHA~\cite{garcia2018first}) in which the hand interacts with the other's hand, respectively. }
\label{fig:frame_annotation}
\end{figure}

As shown in Fig.~\ref{fig:frame_annotation}, we show continuous annotation examples where hand types change over time in the video sequence, following the annotation details above.
In Fig.~\ref{fig:frame_annotation} (a), we provide frames of the action called \textit{``Place Cappuccino''} that interact with the object (cappuccino box). While the left hand keeps the hand on the desk (\textit{``Relaxed Hand''}), the type of the right hand changes as it interacts with the cappuccino box. The right hand type starts as \textit{``Large Diameter''} while holding the box and converts into a \textit{``Fist''} through the intermediate stage of \textit{``Relaxed Hand''} after putting the box down. Further, in Fig.~\ref{fig:frame_annotation} (b), we visualize frames of the action called \textit{``High Five''} that interact with the other's hand. Even though the shape of the hand is the same throughout high-fiving, hand type differs as \textit{``Open Hand''} or \textit{``Dynamic Flatten''} based on whether it interacts with the other person's hand.

\section*{D. Analysis of Datasets and Hand Type Distributions}
\label{sec:d}
\label{sec:dataset}

We train and evaluate overall performance on two landmark datasets (FPHA~\cite{garcia2018first} and H2O~\cite{kwon2021h2o}).

\myparagraph{FPHA} The FPHA~\cite{garcia2018first} (First-Person Hand Action) dataset records 45 action categories from 6 subjects with an Intel RealSence SR300 RGB-D camera mounted on the subject's shoulder. It designs various action scenarios interacting with 26 different objects. Only J=21 joints of the subject's right hand are annotated using 3D poses from the wearable magnetic sensors. It contains 1,175 sequences and separates into 1:1 settings with 600 and 575 videos for the training and evaluation phase in each. Both steps include all subjects and actions.

\myparagraph{H2O} The H2O~\cite{kwon2021h2o} (2 Hands and Objects) dataset captures 4 subjects performing 36 actions, interacting with 8 3D objects. Unlike the FPHA~\cite{garcia2018first} dataset, which only provides labels for the right hand, H2O provides 3D labels for both hands with the total number of $J=21\times{2}$ joints.
It includes 569 videos for the training, including all actions for the first 3 subjects, 122 videos for the validation, and 242 videos of the remaining subject unseen in the training for the test.

\begin{wrapfigure}{r}{0.5\textwidth}
\vspace{-1.2em}
\centering
\includegraphics[width=0.5\textwidth]{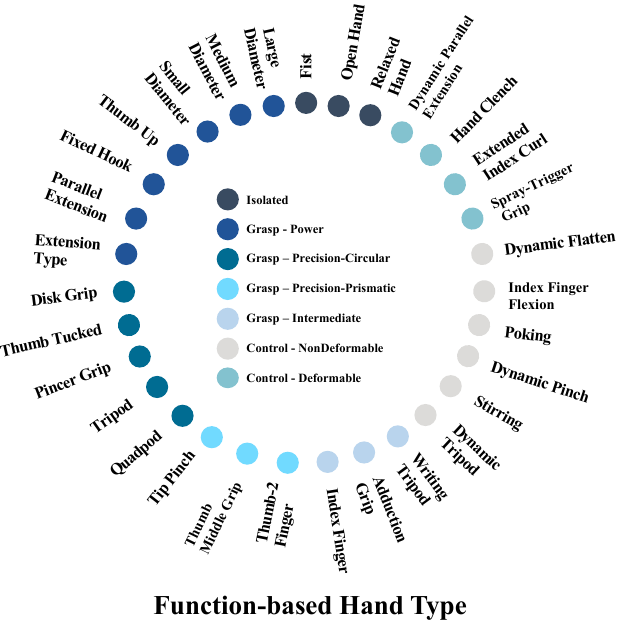}
\vspace{-1.2em}
\caption{
Radial diagram of function-based hand type taxonomy. To explain more real-life hand behaviors, we carefully design a taxonomy based on the functioning perspective.}
\label{fig:radial}
\vspace{-1.2em}
\end{wrapfigure}

\myparagraph{Hand Type Distribution} In this paper, we introduce a newly defined taxonomy of 31 hand types based on functionality. We visualize the distribution of our proposed hand type taxonomy with a radial diagram in Fig.~\ref{fig:radial}. We first categorize the mainly used hand types into \textit{Isolated} and \textit{Interaction} based on whether the hands interact with objects (or other people). We then classify the \textit{Interaction} types into \textit{Grasp} and \textit{Control}, reflecting the role of the hand in the scene. These categories are further subdivided according to the appearance of the hand and the degree of object manipulation. Finally, function-based taxonomy that comprehensively encompasses real-life hand movements allows explaining various hand action scenarios. With our new taxonomy, we annotate hand types across all frames of two datasets.

\begin{figure}[t]
    \centering
    \includegraphics[width=1\textwidth]{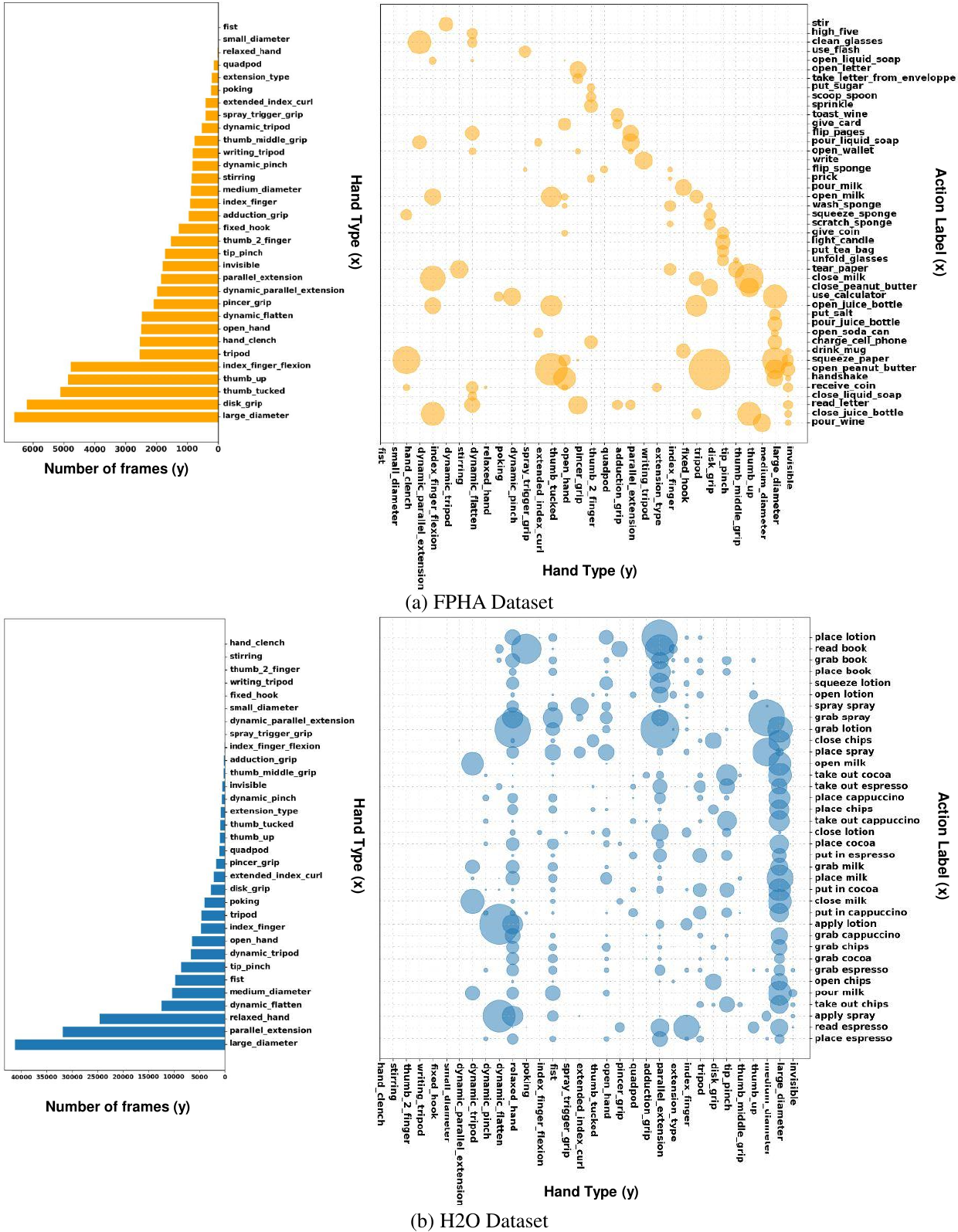}
    \vspace{-1.8em}
    \caption{Hand Type Distributions of the (a) FPHA~\cite{garcia2018first} Dataset (Orange) and (b) H2O~\cite{kwon2021h2o} Dataset (Blue). The histograms (left) show the number of frames per hand type; the x-axis represents the hand type, and the y-axis indicates the number of frames. The scatter plots (right) present the distribution of hand types across each action label; the x-axis represents the action of each dataset, and the y-axis indicates the hand type. The size of the circles illustrates how often each hand type appears within the hand action video sequences.
    }
    \label{fig:distribution}
\end{figure}

To investigate the hand type distributions of both FPHA~\cite{garcia2018first} (Fig.~\ref{fig:distribution} (a)) and H2O~\cite{kwon2021h2o} (Fig.~\ref{fig:distribution} (b)) datasets, we illustrate them via histograms and scatter plots.
We first explore the number of frames per hand type with the histograms (Fig.~\ref{fig:distribution} left) to observe how frequently each hand type occurs within whole hand video sequences. The x-axis represents the hand type, and the y-axis shows the number of frames. As shown in Fig.~\ref{fig:distribution} (left), the distributions of the two datasets exhibit different patterns. In the case of the FPHA, 29 of the 31 hand types are identified, with~\textit{``Large Diameter''},~\textit{``Disk Grip''} and~\textit{``Thumb Tucked''} being the most common. For the H2O, 24 of the 31 hand types are recognized, with~\textit{``Large Diameter''},~\textit{``Relaxed Hand''} and~\textit{``Parallel Extension''} being the most prevalent. We find that the FPHA operates more diverse hand types than H2O for more action scenarios.

\begin{figure}[t]
    \centering
    \includegraphics[width=1\textwidth]{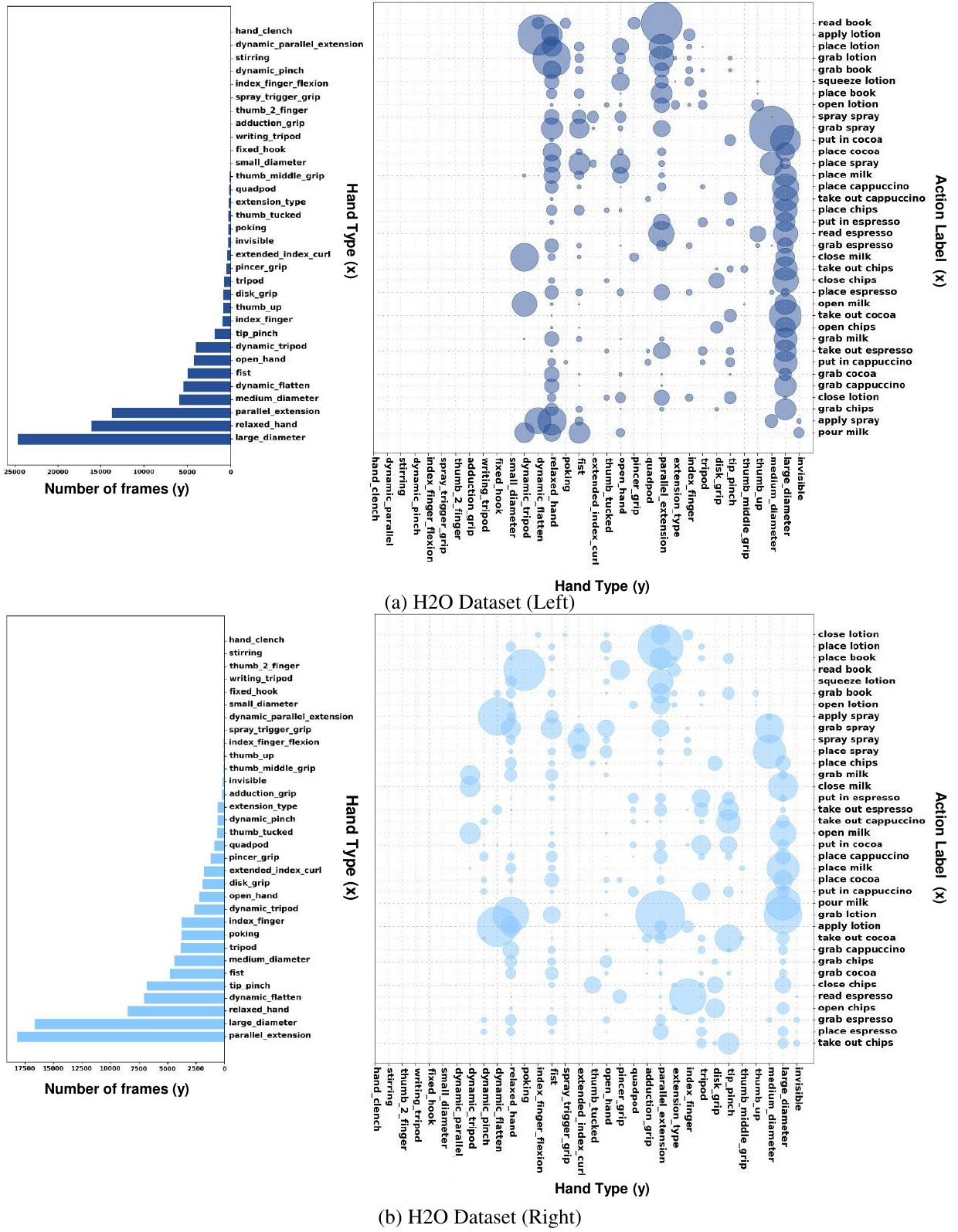}
    \vspace{-1.8em}
    \caption{Hand Type Distributions for the (a) Left hand and (b) Right hand of the H2O~\cite{kwon2021h2o} Dataset. The histograms (left) show the number of frames per hand type; the x-axis represents the hand type, and the y-axis indicates the number of frames. The scatter plots (right) present the distribution of hand types across each action label; the x-axis represents the action of each dataset, and the y-axis indicates the hand type. The size of the circles illustrates how often each hand type appears within the hand action video sequences.
    }
    \label{fig:distribution_h2o}
\end{figure}

Also, we visualize the distribution of hand types across each action label with scatter plots (Fig.~\ref{fig:distribution} right); the x-axis indicates the action of each dataset, and the y-axis represents the hand type. The circle size portrays the appearance frequency of each hand type within each action video clip. As shown in Fig.~\ref{fig:distribution} (right), the FPHA presents monotonous distributions of hand types per action, while the H2O shows a relatively balanced distribution. This difference is because, unlike the H2O, which focuses on both hands, the FPHA annotates only on the right hand and records simple actions repeatedly. Ultimately, FPHA utilizes 29 hand types for 45 action scenarios but is monotonous, and H2O uses 24 hand types for 36 action scenarios but is relatively harmoniously distributed. Note that we further visualize hand type distributions of each left and right hand of the H2O dataset in Fig.~\ref{fig:distribution_h2o}.

\clearpage
\bibliography{egbib}